\documentclass{article}

\PassOptionsToPackage{numbers, compress}{natbib}
 \usepackage[preprint]{neurips_2026}

\usepackage[utf8]{inputenc} 
\usepackage[T1]{fontenc}    
\usepackage{hyperref}       
\usepackage{url}            
\usepackage{booktabs}       
\usepackage{amsfonts}       
\usepackage{nicefrac}       
\usepackage{microtype}      
\usepackage{xcolor}         
\usepackage{graphicx}
\usepackage{wrapfig}
\usepackage{array}
\usepackage{amsmath}
\usepackage{cleveref}
\usepackage{float}
\usepackage{hyperref}
\title{Structured 3D Latents Are Surprisingly Powerful: Unleashing Generalizable Style with 2D Diffusion}

\author{
    Yiran Qiao,
    Yiren Lu,
    Yunlai Zhou,
    Disheng Liu,
    Linlin Hou,
    Rui Yang,
    Yu Yin,
    Jing Ma\thanks{\quad Corresponding author.}
    \\
    Case Western Reserve University \\
    \{yxq350, yxl3538, yxz3057, dxl952, lxh633, rxy337, yxy1421, jxm1384\}@case.edu \\
}

\begin{document}

\maketitle

\begin{abstract}
\looseness -1
3D asset generation plays a pivotal role in fields such as gaming and virtual reality, enabling the rapid synthesis of high-fidelity 3D objects from a single or multiple images. Building on this capability, enabling style-controllable generation naturally emerges as an important and desirable direction. However, existing approaches typically rely on style images that lie within or are similar to the training distribution of 3D generation models. When presented with out-of-distribution (OOD) styles, their performance degrades significantly or even fails. To address this limitation, we introduce \textbf{DiLAST}: 2D \underline{Di}ffusion-based \underline{L}atent \underline{A}wakening for 3D \underline{S}tyle \underline{T}ransfer. Specifically, we leverage a pretrained 2D diffusion model as a teacher to provide rich and generalizable style priors. By aligning rendered views with the target style under diffusion-based guidance, our method optimizes the structured 3D latent representations for stylization. We observe that this limitation stems not from insufficient model capacity, but from the underutilization of structured 3D latents, which are inherently expressive. Despite being trained on comparatively limited data, 3D generation models can leverage 2D diffusion guidance to steer denoising toward specific directions in latent space, thereby producing diverse, OOD styles. Extensive experiments across diverse data and multiple 3D generation backbones demonstrate the effectiveness and plug-and-play nature of our approach. \href{https://yrqiao.github.io/DiLAST/}{\textcolor{blue}{[Project Page]}}
\end{abstract}

\section{Introduction}

The rapid advancement of generative models has become an important component of progress in artificial intelligence (AI), giving rise to the paradigm of AI Generated Content (AIGC). In this field, 2D generative models, with diffusion models \cite{ho2020denoising,song2020denoising,rombach2022high} as a representative example, have achieved remarkable success in synthesizing high-quality and diverse images. Recently, 3D generative models have begun to emerge, aiming to produce realistic and controllable 3D assets for applications such as gaming, virtual reality, and digital content creation \cite{li2024advances}. Early 3D generative models \cite{chen2024comboverse,poole2022dreamfusion,liu2023zero} still leverage the strong generative capabilities of 2D diffusion models, typically through techniques such as score distillation sampling (SDS) or view-conditioned fine-tuning, enabling the generation of 3D assets conditioned on user-provided text or images. More recent efforts have led to the development of true 3D generative models, such as Trellis \cite{xiang2025structured}, UniLat \cite{wu2025unilat3d}, and Hunyuan3D \cite{zhao2025hunyuan3d}. Unlike earlier approaches that rely on 2D diffusion, these methods perform denoising directly in structured 3D latent spaces, from which the final 3D content is decoded. This leads to more faithful modeling of 3D geometry and appearance, marking a significant step toward native 3D generation.

Building on the aforementioned progress, a natural and practically important task is to enable flexible style manipulation of 3D assets, \textit{i.e.}, transferring arbitrary styles onto 3D content, analogous to the well-studied problem of style transfer in 2D \cite{chung2024style,deng2024z,liu2025unziplora,wang2023stylediffusion,xu2025stylessp, zhou2025attention}. Concretely, given a content image and a style image, the goal is to generate a 3D asset whose geometry is consistent with the content image while its appearance reflects the style image. This problem has attracted increasing attention. StyleSculptor \cite{qu2025stylesculptor} enables zero-shot style-guided 3D generation via style-disentangled attention mechanisms, while MorphAny3D \cite{sun2026morphany3d} leverages structured latent representations and attention-based feature blending for high-quality 3D morphing, with extensions to 3D style transfer. However, these methods are largely constrained by the intrinsic capabilities of the underlying 3D generation models. Consequently, the style images are typically required to lie \textit{within or be close to the training distribution}. When given arbitrary out-of-distribution (OOD) styles, their performance degrades significantly or even fails. This limitation significantly restricts their applicability in real-world scenarios, where users often select style images in an unconstrained and arbitrary manner.

\begin{figure}
    \centering
    \includegraphics[width=1\linewidth]{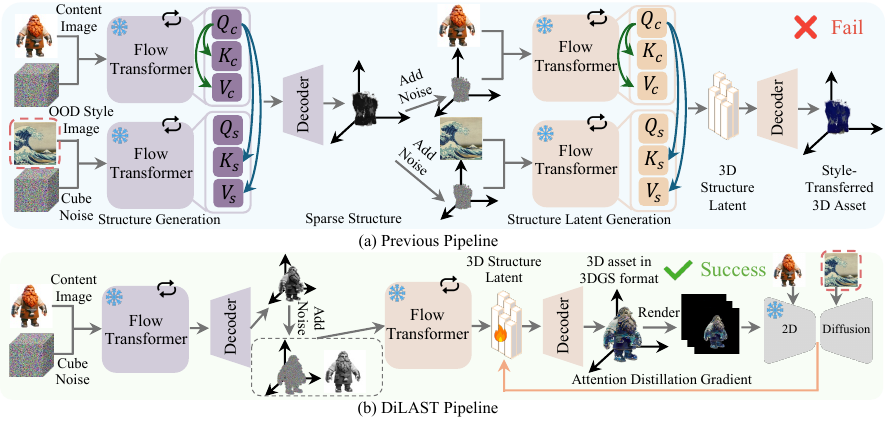}
    \vspace{-5mm}
    \caption{\textbf{Comparison of existing 3D style transfer methods and our DiLAST pipeline.} (a) Existing methods rely solely on \colorbox[HTML]{907D9E}{internal} \colorbox[HTML]{EAD4B8}{attention} of 3D generative models and often \colorbox[HTML]{E35242}{fail} when handling \colorbox[HTML]{DC6C6E}{OOD styles}. (b) In contrast, DiLAST leverages a pretrained \colorbox[HTML]{D9D9D9}{2D diffusion teacher}, whose attention distillation gradients guide the denoising trajectory of 3D latents, \colorbox[HTML]{65AE42}{successfully} transferring arbitrary OOD styles.}
    \vspace{-5mm}
    \label{fig1}
\end{figure}

To tackle this limitation, we propose \textbf{DiLAST}, a framework that leverages 2D diffusion models to awaken the expressive capacity of structured 3D latent representations for style transfer. Specifically, 3D latents can be decoded into various 3D formats. In our implementation, we adopt a 3D Gaussian (3DGS) \cite{kerbl20233d} decoder to produce 3D Gaussian primitives, which are then efficiently rendered into high-fidelity multi-view images. We then feed these rendered images, together with the content and style images, into a pretrained 2D diffusion model. By computing the gradients of the attention distillation loss with respect to the 3D latents, we guide its denoising trajectory, enabling effective style transfer while preserving the underlying content structure. We observe that, although the latent space of 3D generative models is relatively limited, it possesses substantial expressive potential. With appropriate guidance, the denoising process can be steered along specific directions to reach desired stylized latents, leading to the generation of corresponding 3D. Our contributions are summarized as:
\begin{itemize}
    \item We \textbf{identify a key limitation} of existing 3D style transfer methods, namely their reliance on the intrinsic capabilities of underlying 3D generation models, which restricts them to in-distribution styles and leads to poor generalization to OOD inputs.
    \item We propose \textbf{DiLAST, a test-time optimization framework} that leverages pretrained 2D diffusion models to guide the denoising trajectory of structured 3D latents for style transfer, without modifying or retraining the underlying 3D generation models.
    \item We provide the \textbf{key insight} that structured 3D latent spaces are more expressive than commonly assumed, and can be effectively steered via diffusion guidance along specific denoising directions to achieve desired stylizations.
    \item We conduct extensive experiments across diverse data and multiple 3D generation backbones to demonstrate the effectiveness, strong generalization to OOD styles, and \textbf{plug-and-play} nature of our approach across different latent spaces.
\end{itemize}

\section{Related Work}
\subsection{2D \& 3D Generative Model}
2D generative models have evolved through several mainstream paradigms. Early breakthroughs were driven by generative adversarial networks, with GAN \cite{goodfellow2014generative} introducing adversarial learning for realistic image synthesis and StyleGAN \cite{karras2019style,karras2020analyzing} enabling highly controllable and high-fidelity generation. Diffusion models subsequently emerged as the dominant framework, beginning with DDPM \cite{ho2020denoising}, formulating generation as a denoising process, followed by score-based generative models \cite{song2020score}, unifying diffusion and score matching. Building on these foundations, Imagen \cite{saharia2022photorealistic}, Stable Diffusion \cite{rombach2022high}, and SDXL \cite{podell2023sdxl} significantly advanced text-guided image synthesis in both quality and scalability. More recently, flow matching has emerged as an efficient alternative to diffusion, with Flow Matching \cite{lipman2022flow} learning continuous probability flows for generation, and Rectified Flow \cite{liu2022flow} further improving sampling efficiency by producing straighter trajectories.

\looseness -1
Beyond 2D generation, recent advances have also driven rapid progress in 3D generative models. Early works relied on GAN for 3D generation \cite{chan2022efficient,deng2022gram,gao2022get3d,zheng2022sdf}. However, the quality of the generated results was constrained by the representational limitations of GAN. As 2D diffusion models gradually supplanted GANs as the dominant generative paradigm, many 3D generation methods began building upon pretrained diffusion models to inherit their strong generative priors. These methods can be broadly categorized into two groups. The first is represented by DreamFusion \cite{poole2022dreamfusion}, which optimizes 3D assets through SDS using pretrained diffusion models. Many subsequent works \cite{liang2024luciddreamer,tang2023dreamgaussian,lin2023magic3d,tang2023make,wang2023prolificdreamer} further improved the distillation technique built upon this framework. Another line of work \cite{li2023instant3d,liu2023zero,liu2024one,long2024wonder3d,shi2023mvdream} first leverages 2D diffusion models to generate multi-view images, and then reconstructs 3D assets from these generated views. However, these methods often struggle to maintain multi-view consistency, which in turn degrades both geometric fidelity and appearance quality. LRM \cite{hong2023lrm} addresses this issue by directly learning a feed-forward reconstruction model from large-scale data, enabling consistent multi-view generation. Analogous to Stable Diffusion compressing pixel-space diffusion into a latent space relative to DDPM, some works \cite{ntavelis2023autodecoding,chen20253dtopia,lan2025ln3diff++} likewise perform 3D generation in 3D latent spaces, improving training efficiency and generalization capability. Extending these efforts, Trellis \cite{xiang2025structured} introduces a versatile structured 3D latent representations that can be decoded into multiple formats of 3D assets. Such a design is particularly beneficial for downstream tasks, including style transfer.

\subsection{2D \& 3D Style Transfer}
Style transfer is an important downstream task of generative models, and existing 2D style transfer methods can be categorized into several directions. Image-encoder–based methods \cite{ye2023ip,qi2024deadiff,li2024styletokenizer,lei2025stylestudio} learn explicit style representations from reference images using dedicated image encoders, and combine these representations with text or image conditions to guide stylized generation. Image-driven style transfer methods \cite{wang2023stylediffusion,deng2024z,jones2024customizing,wang2025sigstyle,wang2025omnistyle} instead use a reference image as the style source directly, transferring visual appearance through feature matching or attention-based guidance. Textual-inversion–based methods \cite{gal2022image,zhang2023inversion} encode a specific style into newly learned text tokens, allowing the style to be invoked through the text encoder without modifying the generative backbone. LoRA-based methods \cite{shah2024ziplora,liu2025unziplora,frenkel2024implicit,yang2025qr} inject style through adapting a small set of low-rank weights, storing style-specific knowledge in lightweight model parameters rather than prompts or reference features.

\looseness -1
Similarly, several works have extended style transfer to the 3D domain. NeRF \cite{mildenhall2021nerf} is a widely used 3D representation, and several works \cite{liu2023stylerf,fan2022unified} perform style transfer by transferring color and appearance patterns from reference images onto NeRF-based 3D scenes. More recently, 3DGS has gradually replaced NeRF as a dominant paradigm for 3D representation. StyleGaussian \cite{liu2024stylegaussian} embeds and transforms 2D style features within 3D Gaussians, enabling real-time stylization through a dedicated feature decoding pipeline. StyleSplat \cite{jain2024stylesplat} enables localized 3D object stylization by selectively fine-tuning segmented Gaussians under reference-style guidance. Another direction focuses on transferring texture styles onto given 3D shapes, represented by StyleTex \cite{xie2024styletex} and TEXTure \cite{richardson2023texture}. Leveraging the strong capabilities of recent 3D generative models, StyleSculptor \cite{qu2025stylesculptor} enables zero-shot style-controllable 3D generation through disentangled attention for texture and geometry dual guidance. MorphAny3D \cite{sun2026morphany3d} performs training-free 3D style transfer by blending structured latent features through tailored attention mechanisms within the structure latents. However, both of these recent 3D style transfer methods remain limited in handling OOD styles, a challenge our work is designed to address.

\begin{figure}
    \centering
    \includegraphics[width=1\linewidth]{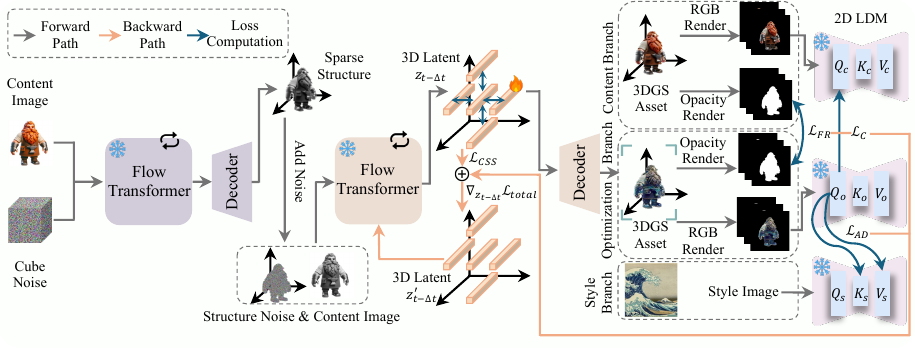}
    \looseness -1
    \caption{Overview of our DiLAST. We leverage a 2D LDM as a teacher and optimize the 3D latent through three parallel branches, guiding the generated 3D asset toward the target style. The 3D asset highlighted by the green bracket, generated at the final optimization step, is the final output.}
    \vspace{-3mm}
    \label{fig2}
\end{figure}

\section{Method}
\vspace{-1mm}
\subsection{Preliminaries}
\looseness -1
\textbf{Latent Diffusion Model (LDM),} with Stable Diffusion (SD) \cite{rombach2022high} being the most representative example, has achieved state-of-the-art performance in 2D image and video generation. LDM incorporates cross-attention mechanisms to inject conditioning information, extending diffusion models into conditional generators. Specifically, an input image $x$ is first encoded into a latent space through a pretrained VAE $\mathcal{E(\cdot)}$. A UNet denoising backbone $\epsilon_\theta(\cdot)$ is then trained to predict the injected noise during the diffusion process by minimizing the mean squared error between the predicted noise and the added noise $\epsilon$:
\begin{equation}
    \mathcal{L}_{\mathrm{LDM}}
:= 
\mathbb{E}_{\mathcal{E}(x),\, y,\, \epsilon \sim \mathcal{N}(0,1),\, t}
\left[
\left\|
\epsilon - \epsilon_{\theta}(w_t, t, \tau_{\theta}(y))
\right\|_2^2
\right],
\end{equation}
where $y$ denotes the condition and $t$ represents the timestep, $w_t$ is the noisy latent at $t$, $\tau_{\theta}(\cdot)$ is an encoder that  projects $y$ to an intermediate representation. This conditioning representation is then embedded into intermediate UNet layers through cross-attention mechanisms:
\begin{equation}
    \mathrm{Attention}(Q,K,V)
=
\mathrm{softmax}\left(
\frac{QK^{T}}{\sqrt{d}}
\right)V,
\end{equation}
where $Q=W_Q^{(i)}\varphi_i(z_t),\; K=W_K^{(i)}\tau_\theta(y),\; V=W_V^{(i)}\tau_\theta(y)$. $i$ indexes the $i$-th cross-attention layer in the UNet, and $\varphi_i(z_t)$ denotes the intermediate latent feature at that layer derived from the noisy latent $z_t$. Note that for self-attention in LDM, the query, key, and value are all projected from the same latent feature, \textit{i.e.}, $Q=W_Q^{(i)}\varphi_i(z_t)$, $K=W_K^{(i)}\varphi_i(z_t)$, and $V=W_V^{(i)}\varphi_i(z_t)$.

\textbf{Trellis} \cite{xiang2025structured} is a foundation model for 3D generation that can synthesize high-quality 3D assets in multiple formats (\textit{e.g.} 3DGS, NeRF, Mesh) conditioned on user-provided text or images. It adopts a two-stage generation framework based on flow models. In the first stage, DINOv2 is first used to extract image features $f$, which serve as conditioning for both stages of generation. An initialized 3D latent noise grid $s_t$ is then fed into a flow transformer $v_s(\cdot)$ to predict the velocity field:
\begin{equation}
    s_{t-\Delta t}
=
s_t
-
v_{s}(f,s_t,t)\Delta t,
\end{equation}
where $\Delta t$ is the step length. At the final denoising step, the resulting $s_0$ is decoded into a sparse structure represented by active voxels. Each active voxel is injected with an initialized latent noise vector $z_t$, which is then passed into the flow transformer $v_l(\cdot)$ to predict the velocity field in the second stage:
\begin{equation}
    z_{t-\Delta t}
=
z_t
-
v_{l}(f,z_t,t)\Delta t,
\end{equation}
Finally, the resulting $z_0$ is decoded into the desired 3D asset in the target output format.

\subsection{2D Diffusion-Guided Test-Time Optimization}
Given a content image and a style image, our goal is to generate a 3D asset conditioned on the content image, while ensuring that its style is consistent with the style image. Previous works typically rely on a pretrained 3D generative model with separate content and style branches, where stylization is achieved through attention injection or feature recombination within the model. However, current 3D generative models are primarily designed for object-level generation. When arbitrary OOD style images are used to condition the style branch, the resulting distribution mismatch often causes misaligned query-key-value associations in attention layers, leading to unreliable style guidance and poor stylized 3D assets.

In contrast, 2D latent diffusion models are far less susceptible to this issue, as they are trained on much larger-scale and more diverse data, endowing them with stronger robustness to distribution shifts and more reliable query-key-value correspondences in attention layers. However, since the latent spaces of 2D diffusion and 3D generative models are not aligned, directly performing attention injection across them is infeasible. Motivated by this, we adopt the \textit{attention distillation} mechanism, shifting style guidance from the 3D generator to a pretrained 2D latent diffusion teacher, thereby confining all attention interactions to the 2D LDM.

\looseness -1
Since 2D latent diffusion models take images as input, they naturally serve as a reliable bridge between 2D and 3D models. Leveraging the versatile structured 3D latents provided by Trellis, we adopt the 3DGS decoder to obtain Gaussian primitives representing the generated 3D asset, due to its favorable properties, including efficient rendering, high-fidelity synthesis, and differentiable rasterization:
\begin{equation}
    \mathcal{D}_{GS} :
\left\{
(z_i,p_i)
\right\}_{i=1}^{L}
\mapsto
\left\{
\left\{
(o_i^{k},c_i^{k},s_i^{k},\alpha_i^{k},r_i^{k})
\right\}_{k=1}^{K}
\right\}_{i=1}^{L},
\end{equation}
where $\mathcal{D}_{GS}$ denotes the 3D Gaussian decoder, $p_i$ represents the $i$-th active voxel, $z_i$ is the latent vector associated with that voxel, and $L$ is the number of active voxels. Each active voxel is decoded into $K$ Gaussians, parameterized by $(o_i^k,c_i^k,s_i^k,\alpha_i^k,r_i^k)$, where $o_i^k$ denotes position, $c_i^k$ color, $s_i^k$ scale, $\alpha_i^k$ opacity, and $r_i^k$ rotation. Then, given a specific camera pose $\pi$, the color $I_\pi(x,y)$ at each pixel $(x,y)$ in the rendered 2D image is computed through alpha-blending over the set of Gaussians $\mathcal{N}$ that are visible under the view and contribute to the pixel due to spatial overlap:
\begin{equation}
        I_\pi(x,y) = \sum_{u \in \mathcal{N}} c_u \alpha_u \prod_{v=1}^{u-1} (1 - \alpha_v).
\end{equation}
After obtaining the rendered images, we initialize three parallel branches in the same 2D LDM. The first is a \textit{content branch}, whose input consists of rendered views from a 3D asset generated by Trellis in 3DGS format, conditioned on the content image. The second is a \textit{style branch}, which directly takes the style image as input. The third is an \textit{optimization branch}, whose input consists of rendered views decoded from the 3D asset at a given timestep during optimization. For this branch, we preserve the first-stage sparse structure generated by Trellis under the content image condition. In the second stage, we instead use the grayscale content image as conditioning to reduce structural drift while preventing the original content style from biasing the learning of the target style. This design is well motivated, as style transfer should not induce large structural changes to the underlying 3D asset. Moreover, stylization in the 3DGS representation modifies Gaussian parameters during optimization, which is sufficiently expressive to capture both geometric and appearance-level style variations. With the three branches defined, the attention distillation loss can be formulated as follows:
\begin{equation}
    \mathcal{L}_{AD}
=
\left\|
\mathrm{SelfAttn}(Q_o,K_o,V_o)
-
\mathrm{SelfAttn}(Q_o,K_s,V_s)
\right\|_1,
\end{equation}
where $Q_o$, $K_o$, and $V_o$ denote the query, key, and value from the optimization branch, while $K_s$ and $V_s$ denote the key and value from the style branch. At a high level, this objective encourages the optimization branch to produce attention responses that match those obtained when querying the style branch. Intuitively, once querying the style branch yields the same response as querying itself, the optimization branch has acquired the desired style characteristics. For content preservation, we adopt the content loss proposed in \cite{zhou2025attention}, defined as follows:
\begin{equation}
    \mathcal{L}_{C}
=
\left\|
Q_o-Q_c
\right\|_1,
\end{equation}
where $Q_c$ is the query from the content branch. With the attention distillation loss and content preservation loss defined, we compute the gradient of the total loss with respect to the 3D latents and use it to steer the optimization trajectory during denoising:
\begin{equation}
    z_{t-\Delta t}:=z_{t-\Delta t}-\eta\nabla_{z_{t-\Delta t}}\mathcal{L}_{total},
\end{equation}
where $\eta$ controls the guidance strength. $\mathcal{L}_{total}=\mathcal{L}_{AD}+\lambda\mathcal{L}_C$, where $\lambda$ is the weight of content loss. More specifically, we do not modify the parameters of the flow transformer, since this would complicate optimization and potentially degrade velocity estimation. Instead, the transformer first predicts a standard flow update from $z_t$ to $z_{t-\Delta t}$, after which the gradient of the objective is used to steer the direction of this update. The guided latent is then fed back into the flow transformer for the next prediction, iteratively refining the denoising trajectory until the final latent $z_0$ is reached.

\subsection{Floater Removal}
\looseness -1
The Gaussian parameters are decoded from the optimized 3D latent representations. During style-driven optimization, without additional regularization, some parameters, including positions, scales, and rotations, may undergo excessive deviations. In 3D space, these deviations manifest as floating Gaussians (floaters), producing artifacts that degrade the quality of the generated 3D asset. Inspired by \cite{hakie2025fix,rogge2025object}, we introduce a floater removal loss to regularize such undesirable deviations:
\begin{equation}
    \mathcal{L}_{FR}
=
\frac{1}{NHW}
\sum_{n=1}^{N}
\sum_{h=1}^{H}
\sum_{w=1}^{W}
\left\|
A^{(n)}(h,w)-M^{(n)}(h,w)
\right\|_1,
\end{equation}
where $N$ denotes the number of rendered views, and $H$ and $W$ denote the height and width of each image, respectively. $A$ denotes the alpha compositing map rendered from the currently optimized 3D asset, where each pixel accumulates the opacity contributions of visible Gaussians. $M$ denotes the corresponding reference alpha map rendered from the content-conditioned 3D asset, serving as a silhouette mask.

\subsection{Color Speckle Suppression}
\looseness -1
During style-driven optimization, without proper regularization, Gaussian color parameters may drift beyond a stable range, after which they become clipped by the rendering process. In RGB space, this often manifests as saturated extreme values (\textit{e.g.}, near 0 or 255 in individual channels), producing highly saturated spurious color patches that degrade visual quality. Inspired by \cite{fridovich2022plenoxels}, we introduce a color speckle suppression loss to regularize such undesirable color deviations:
\begin{equation}
\mathcal{L}_{\mathrm{CSS}}
=
\frac{1}{|\Gamma|}
\sum_{d\in \Gamma}
\frac{1}{|\Omega_d|}
\sum_{(i,j)\in \Omega_d}
\sqrt{
\|z_i-z_j\|_2^2,
}
\end{equation}
where $\Gamma \subseteq \{l,m,n\}$ denotes the set of valid spatial directions containing at least one neighboring voxel pair, and $|\Gamma|$ is the number of such directions. For each direction $d$, $\Omega_d$ denotes the set of adjacent active voxel pairs along direction $d$, with cardinality $|\Omega_d|$. Specifically, for voxel coordinates $p_i=(l_i,m_i,n_i)$, a pair $(i,j)\in\Omega_d$ satisfies $p_j=p_i+\delta_d$, where $\delta_l=(1,0,0),\delta_m=(0,1,0),\delta_n=(0,0,1)$. This loss promotes local smoothness in the structured 3D latent field by penalizing large discrepancies between neighboring voxels, thereby discouraging abrupt variations in latent features and suppressing spurious artifacts.

\subsection{Overall Guidance}
Based on the preceding subsections, the total loss for optimizing the 3D latents is redefined as follows:
\begin{equation}
    \mathcal{L}_{total}=\mathcal{L}_{AD}+\lambda_{1}\mathcal{L}_C+\lambda_{2}\mathcal{L}_{FR}+\lambda_{3}\mathcal{L}_{CSS},
\end{equation}
\looseness -1
where $\lambda_1$, $\lambda_2$, and $\lambda_3$ are the weighting coefficients associated with the corresponding loss terms, respectively. Note that Trellis is used here only as an illustrative example to present our framework, rather than a restriction of its applicability. Our method is inherently plug-and-play and can be integrated with other 3D generation backbones, as will be validated in the experimental section.

\section{Experiments}
\begin{figure}
    \centering
    \includegraphics[width=0.96\linewidth, height=5.5in]{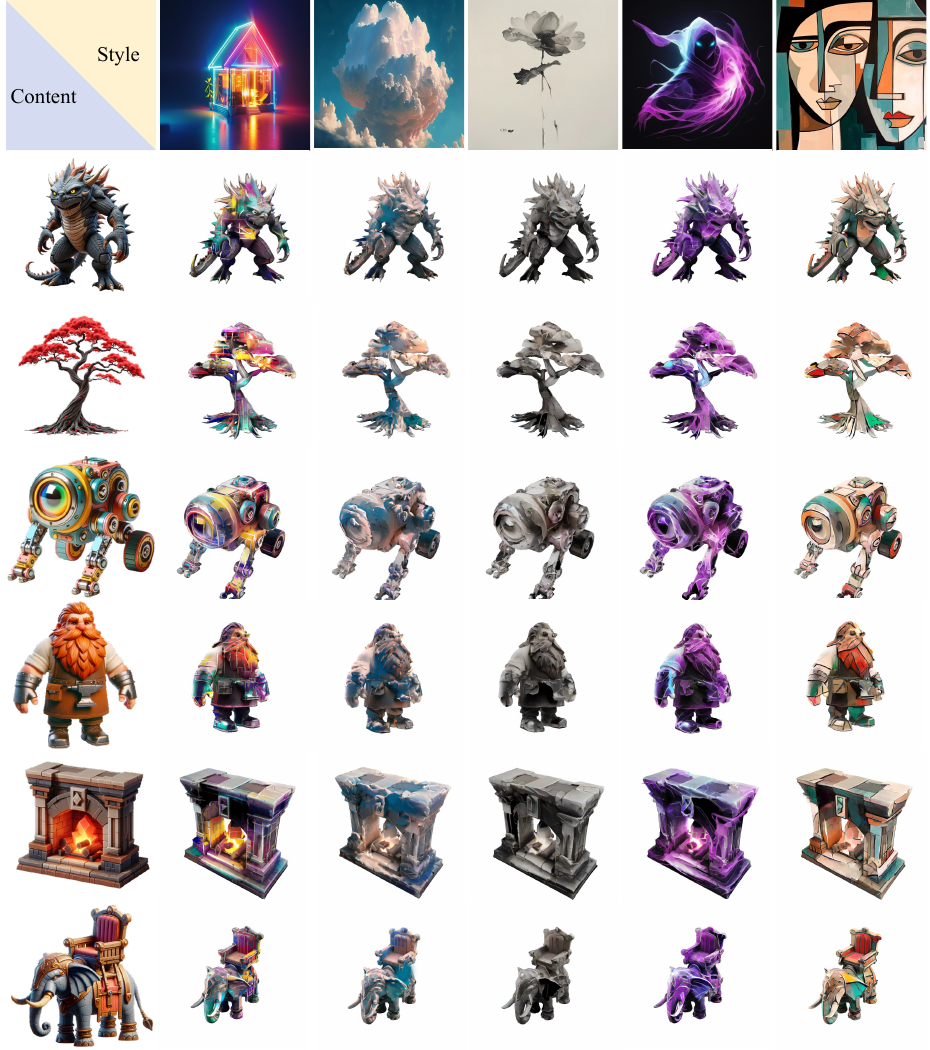}
    \vspace{-1mm}
    \caption{Qualitative results of style transfer using our method.}
    \vspace{-4mm}
    \label{fig3_1}
\end{figure}
\begin{figure}
    \centering
    \includegraphics[width=1\linewidth,height=5.6in]{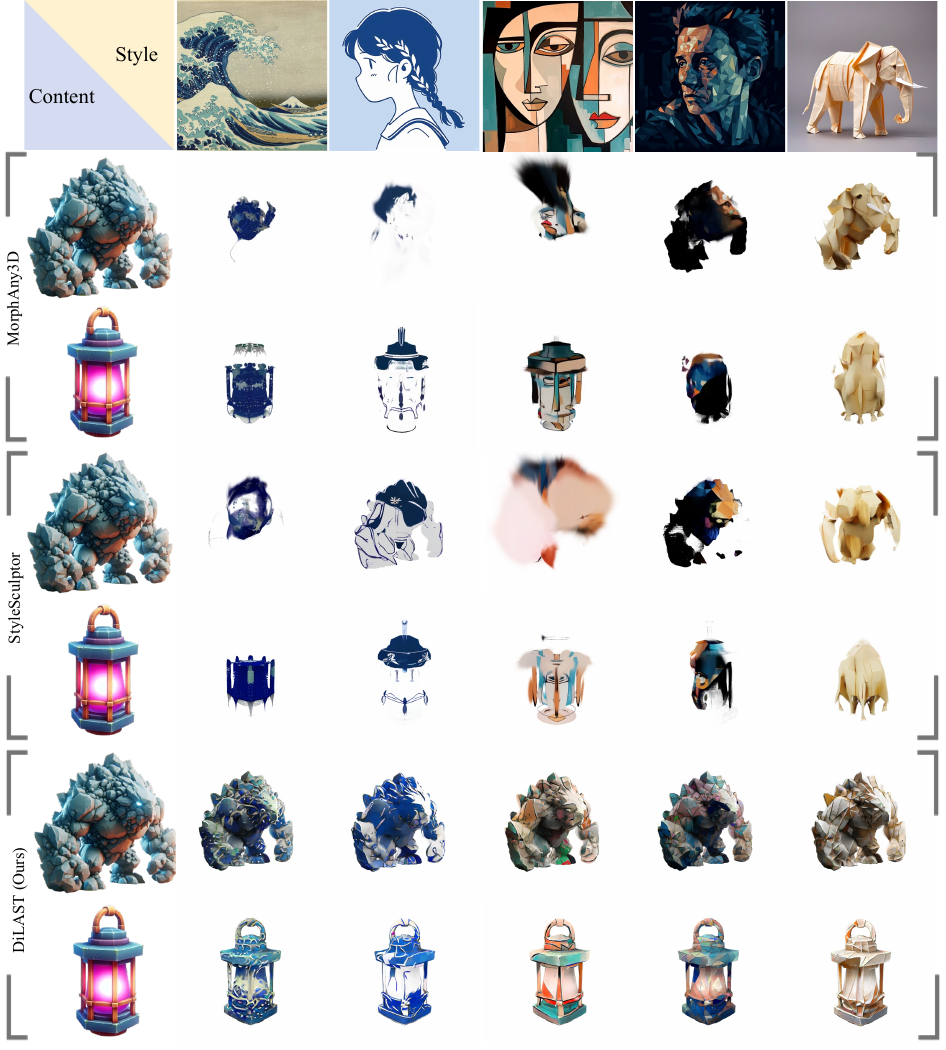}
     \vspace{-1mm}
    \caption{Qualitative comparison of style transfer results across different methods.}
    \vspace{-4mm}
    \label{fig4}
\end{figure}
In this section, we conduct extensive experiments to answer the following research questions: \textbf{RQ1:} Can DiLAST achieve high-quality 3D style transfer across diverse styles? \textbf{RQ2:} How does DiLAST compare with existing methods? \textbf{RQ3:} What is the contribution of each loss component (ablation study)? \textbf{RQ4:} Does our method generalize effectively to other 3D generation backbones?

\subsection{Experiment Settings}
\vspace{-1mm}
\textbf{Datasets \& Baselines.} For evaluation, we sample content images from the TRELLIS-500K \cite{xiang2025structured} dataset and style images from InstantStyle-Plus \cite{wang2024instantstyle}. Since style transfer in structured 3D latent based 3D generative models remains relatively underexplored, we select two representative state-of-the-art methods, MorphAny3D \cite{sun2026morphany3d} and StyleSculptor \cite{qu2025stylesculptor}, as baselines.

\looseness -1
\textbf{Metric.} We first adopt HPSv3 (Human Preference Score v3) \cite{ma2025hpsv3}, a learned metric that predicts human preference scores by aligning model outputs with human aesthetic and semantic judgments. As HPSv3 requires a text prompt, we design two types of prompts for evaluation. The first is a generic prompt (“a stylized object”), which minimizes potential bias introduced by prompt semantics. The second is a specific prompt: for each style image, we use a vision-language model (VLM) \cite{openai2026chatgpt55} to generate a textual description of its style in the form of “<style name> style.” Based on this description, we then construct the final prompt as “a <object name> in <style name> style.” This design allows us to better assess style consistency between the generated 3D assets and the target styles. We further introduce an aesthetic score by leveraging VLMs to assess the visual quality of rendered multi-view images from stylized 3D assets across multiple perceptual dimensions. To improve reliability, we incorporate both closed-source \cite{openai2026chatgpt55} and open-source \cite{bai2025qwen3} VLMs. We also conducted a user preference study with 10 volunteers using the same evaluation criteria. Detailed prompts can be found in Appendix \ref{Appendix A}.

\looseness -1
\textbf{Implementation Details.} All experiments are conducted on NVIDIA A100 GPUs. We use TRELLIS-image-large \cite{xiang2025structured} as the 3D generation model and Stable Diffusion v1-5 as the 2D latent diffusion model. All model parameters are frozen during optimization. For the 2D LDM, we use an empty text condition, since our framework does not rely on text prompts. Unless otherwise specified, we set the loss weights to $\lambda_1=0.2$, $\lambda_2=20$, and $\lambda_3=10$. The number of sampling steps for the flow transformer is set to 500 to allow fine-grained control over the denoising trajectory, and the guidance strength $\eta$ is set to 5. To reduce memory consumption and improve efficiency, we use rendered images from 5 fixed viewpoints at each optimization step. 
More implementation details are provided in Appendix \ref{Appendix B}.

\begin{wraptable}{r}{0.65\textwidth}
\centering
\looseness -1
\vspace{-6mm}
\caption{Quantitative comparison. The best results are in \textbf{bold}. GP: general prompt, SP: specific prompt, AS denotes aesthetic score (averaged on two VLMs), and UP denotes user preference.}
\vspace{1mm}
\label{tab:quantitative}
\resizebox{0.65\textwidth}{!}{
\begin{tabular}{ccccc}
\hline
Method        & HPSv3 (GP)$\uparrow$ & HPSv3 (SP)$\uparrow$ & AS (\%)$\uparrow$ & UP (\%)$\uparrow$ \\ \hline
MorphAny3D    & -4.5784    & -4.7300    & 1.43         & 0.00    \\
StyleSculptor & 0.7625     & -0.7858    & 14.29         & 10.00   \\
\textbf{DiLAST (Ours)} & \textbf{3.0264}     & \textbf{3.3314}     &  \textbf{84.28}       & \textbf{90.00}   \\ \hline
\end{tabular}
}
\end{wraptable}
\subsection{Experiment Results}
\looseness -1
\textbf{Effectiveness of DiLAST (RQ1).} We present the qualitative results in Figure \ref{fig3_1}. Due to space limitations, we only show a subset of content-style combinations. More results are provided in Appendix \ref{Appendix C}. As shown in the figure, DiLAST successfully transfers diverse styles to multiple 3D assets while preserving the underlying content structure. Since we use 3DGS as the 3D asset representation, the textures and brushstroke patterns of the style images can still be effectively reflected even without modifying the first-stage sparse structure. This is because changes in the optimized 3D latents induce corresponding updates to the parameters of the 3D Gaussian primitives.

\begin{wrapfigure}{r}{0.45\linewidth}
\vspace{-3mm}
    \centering    \includegraphics[width=\linewidth]{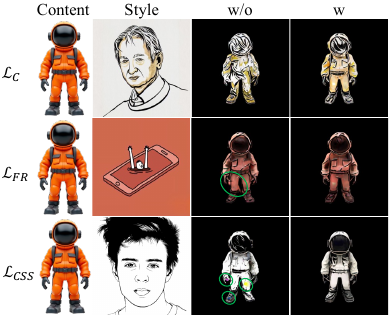}

    \caption{Contribution of Each Loss Component. “w/o” denotes removing the corresponding loss term, while “w” denotes including it.}
    \vspace{-3mm}

    \label{fig5}
\end{wrapfigure}
\looseness -1
\textbf{Comparison between DiLAST and Baseline Methods (RQ2).} We comprehensively compare DiLAST with the baseline methods through both quantitative and qualitative results. Table \ref{tab:quantitative} reports the quantitative results, where our method outperforms the baselines across all metrics, especially on AS and UP, which provide more comprehensive evaluations. For HPSv3, our method achieves the best performance under both the generic and specific prompts. Notably, when switching from the generic prompt to the specific prompt, only our method shows an improvement, indicating that DiLAST receives higher scores under more fine-grained prompts. This further demonstrates that the stylized 3D assets generated by our method achieve the best balance between content preservation, reflected by the object description in the prompt, and style transfer, reflected by the style description in the prompt. Figure \ref{fig4} shows the qualitative results. As can be observed, the stylized 3D assets generated by the baseline methods contain noticeable artifacts or even fail completely. In contrast, our method effectively handles OOD-style images and achieves a better balance between content preservation and style transfer.

\begin{wrapfigure}{r}{0.45\linewidth}
    \centering
    \includegraphics[width=\linewidth]{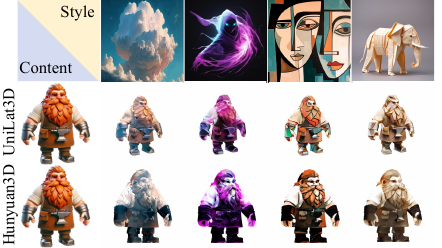}

    \caption{Qualitative results by using other 3D generative models.}
    \vspace{-1mm}
    \label{fig6}
\end{wrapfigure}
\textbf{Ablation Study (RQ3).} Figure \ref{fig5} illustrates the contribution of each loss term. The first row shows the effect of $\mathcal{L}_C$. Without $\mathcal{L}_C$, severe content leakage from the style image is observed, whereas adding this loss effectively alleviates the issue. The second row shows the effect of $\mathcal{L}_{FR}$. Without $\mathcal{L}_{FR}$, floating Gaussians appear in the regions highlighted by the green circles; after adding this constraint, these floaters are suppressed. The third row shows the effect of $\mathcal{L}_{CSS}$. Without $\mathcal{L}_{CSS}$, especially when using light-colored style images, many color artifacts appear on the generated 3D assets. The RGB values of the artifacts highlighted by the green circles are (255,0,255) and (255,255,0), indicating clear overflow and clipping behavior in the 3DGS decoder. Adding this loss term mitigates these artifacts.

\looseness -1
\textbf{Plug-and-Play (RQ4).} In this paper, we mainly use Trellis as the 3D generation model. However, this does not mean that our method is limited to Trellis or cannot be adapted to other 3D latent based 3D generation models. In this section, we select two representative 3D generation models, UniLat3D \cite{wu2025unilat3d} and Hunyuan3D \cite{zhao2025hunyuan3d}, and combine them with our method to generate stylized 3D assets. Unlike Trellis, UniLat3D adopts a single-stage generation process, while Hunyuan3D first decodes the 3D latents into multi-view images and then bakes them onto the mesh structure generated in the first stage. As shown in Figure \ref{fig6}, our method exhibits a strong plug-and-play property. This demonstrates not only the flexibility of our framework, but also that 3D latents in other latent spaces likewise possess strong potential for stylization.

\vspace{-2mm}
\section{Conclusion}
\vspace{-2mm}
\looseness -1
In this paper, we propose DiLAST, a test-time optimization framework that leverages 2D diffusion guidance to awaken structured 3D latents for generalizable style transfer. By steering the denoising trajectory with attention distillation, DiLAST transfers out-of-distribution styles while preserving content structure. We highlight the untapped potential of 3D latent spaces and provide a flexible solution for OOD 3D stylization. In the future, we hope to extend this framework to broader 3D editing tasks beyond style transfer.
\clearpage
\bibliographystyle{unsrtnat}
\bibliography{ref}


\clearpage
\appendix
\begin{center}
    {\Large\bfseries Appendix}
\end{center}
\section{Additional Metric Details}
\label{Appendix A}
\begin{table}[ht]
\centering
\caption{Style images and corresponding style names.}
\begin{tabular}{
>{\centering\arraybackslash}m{0.18\linewidth}
>{\centering\arraybackslash}m{0.65\linewidth}
}
\hline
Style Image & Style Name \\
\hline
\includegraphics[width=0.5\linewidth]{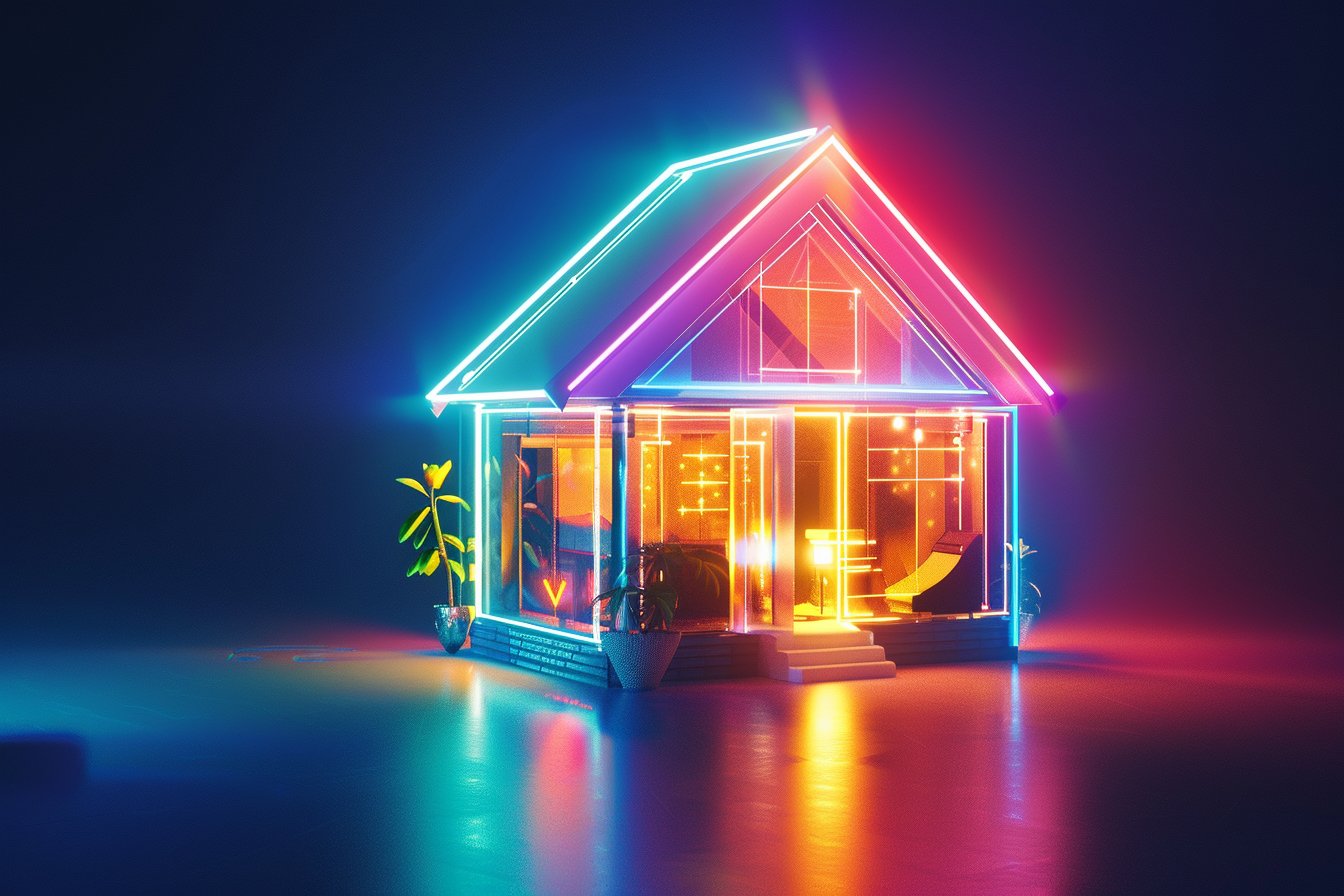}  & Neon cyberpunk glow style \\
\includegraphics[width=0.5\linewidth]{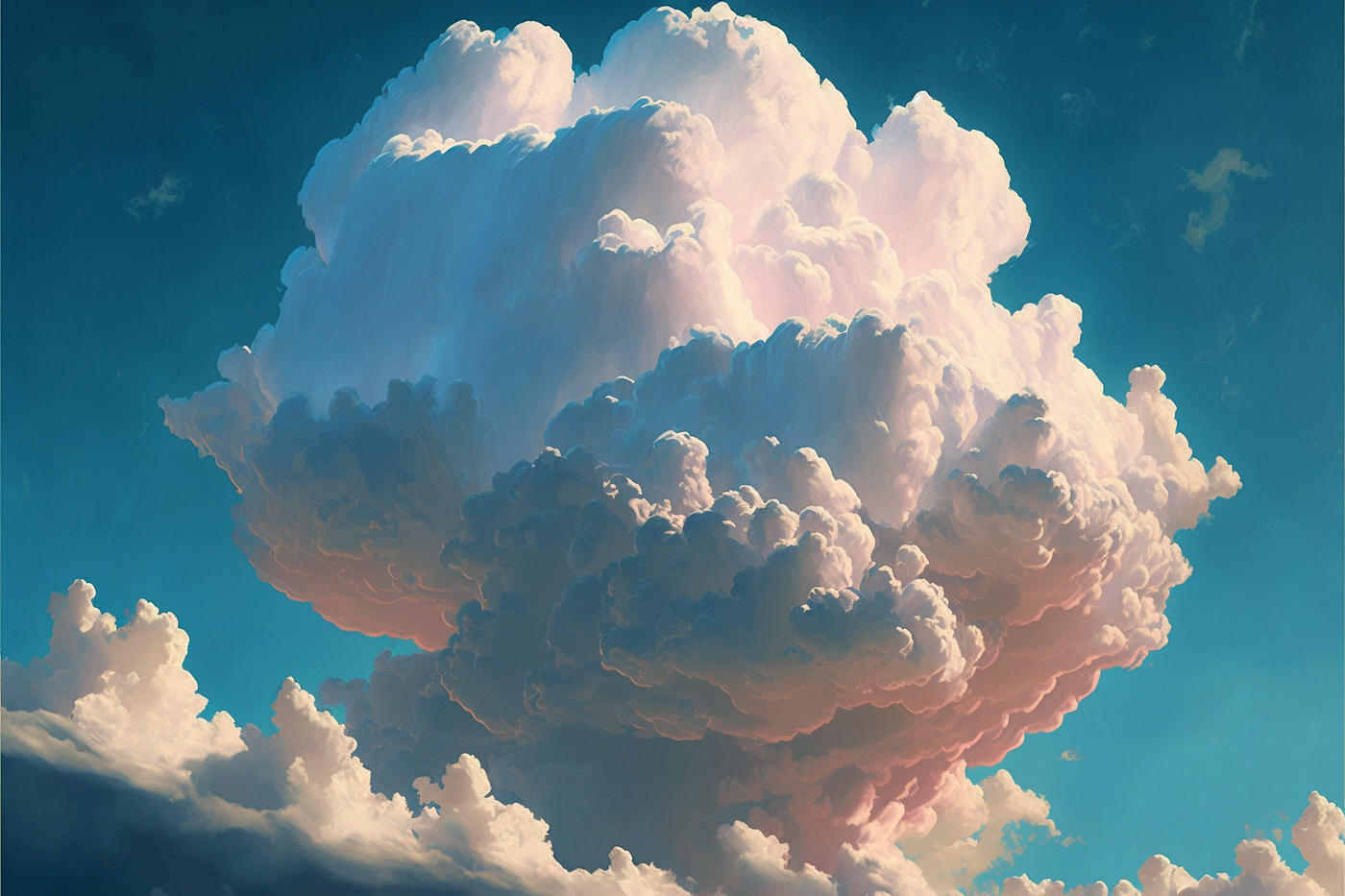}  & Soft atmospheric cloudscape painting style \\
\includegraphics[width=0.5\linewidth]{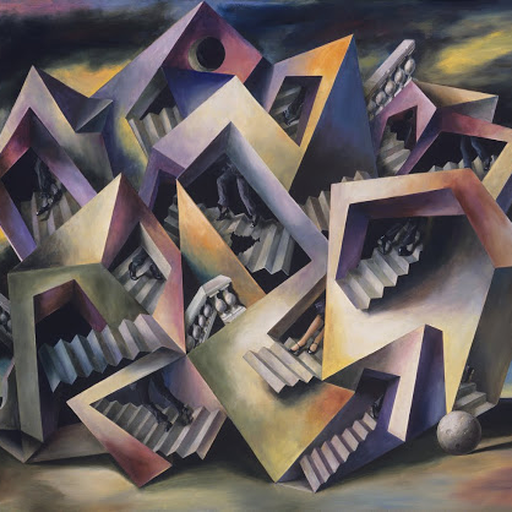}  & Geometric cubist surrealism style \\
\includegraphics[width=0.5\linewidth]{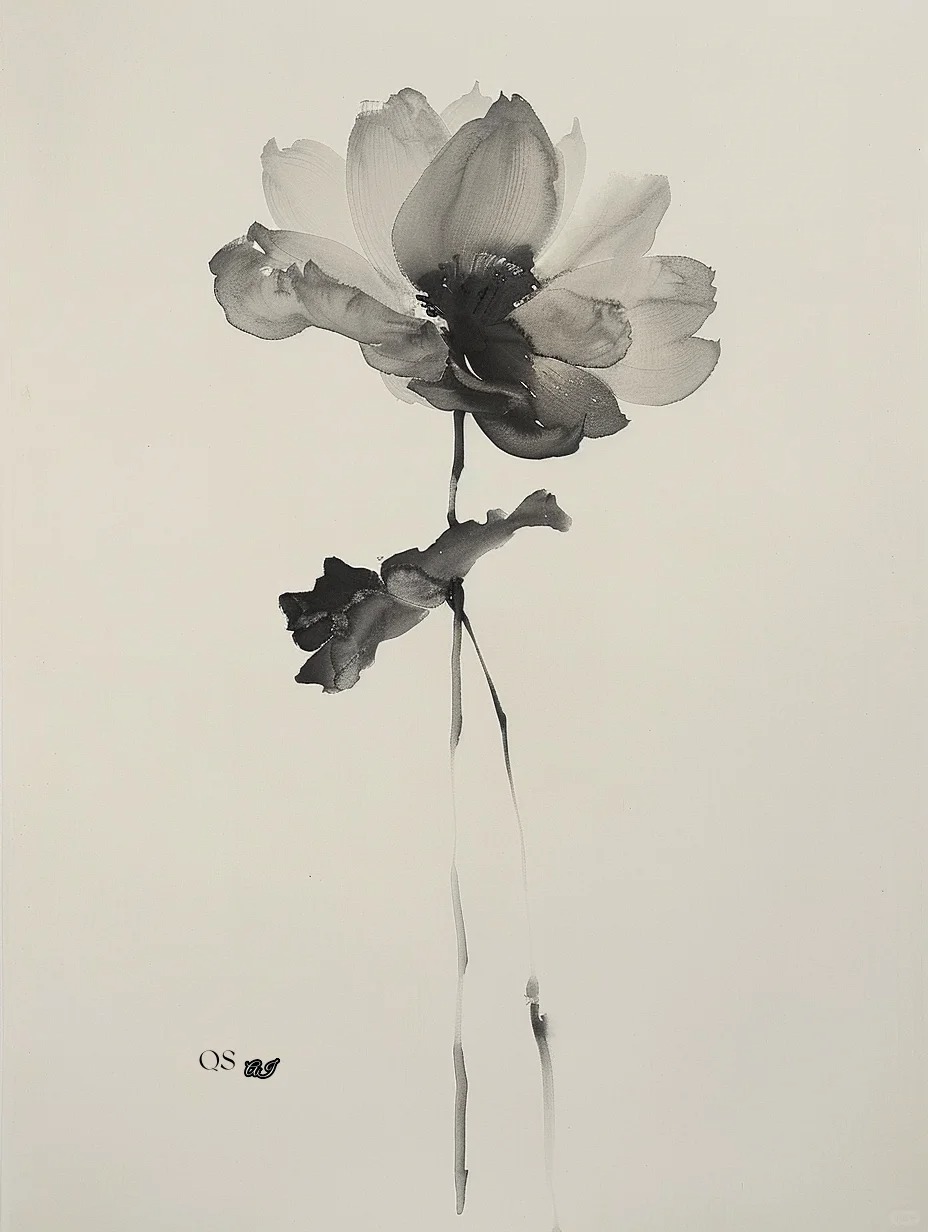}  & Monochrome ink wash minimalist style \\
\includegraphics[width=0.5\linewidth]{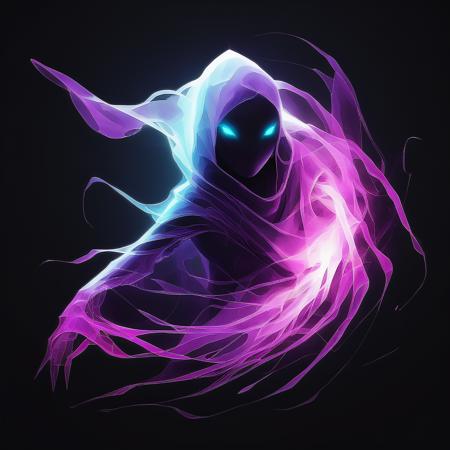}  & Ethereal neon fantasy illustration style \\
\includegraphics[width=0.5\linewidth]{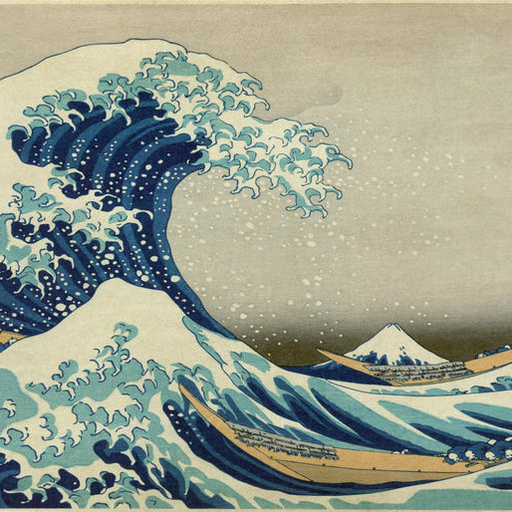}  & Japanese ukiyo-e woodblock print style \\
\includegraphics[width=0.5\linewidth]{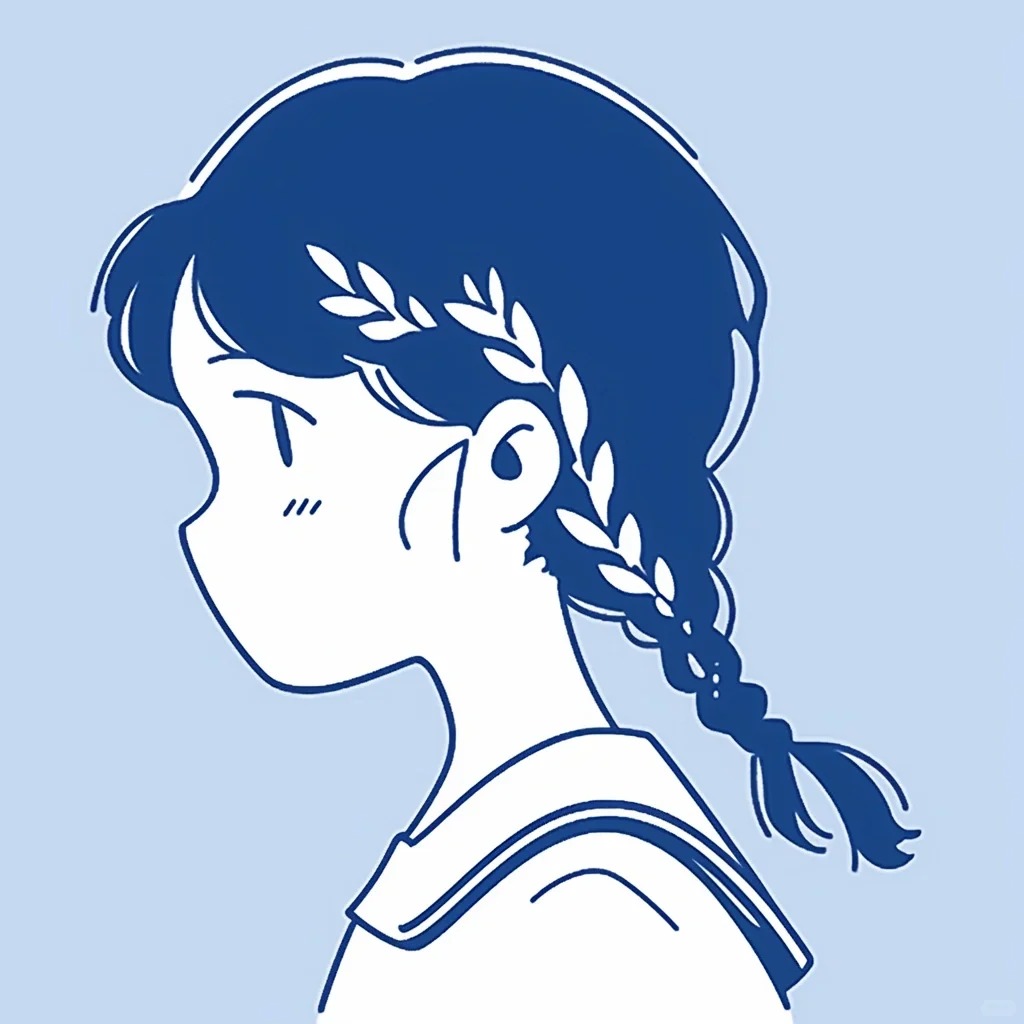}  & Minimal blue line-art anime style \\
\includegraphics[width=0.5\linewidth]{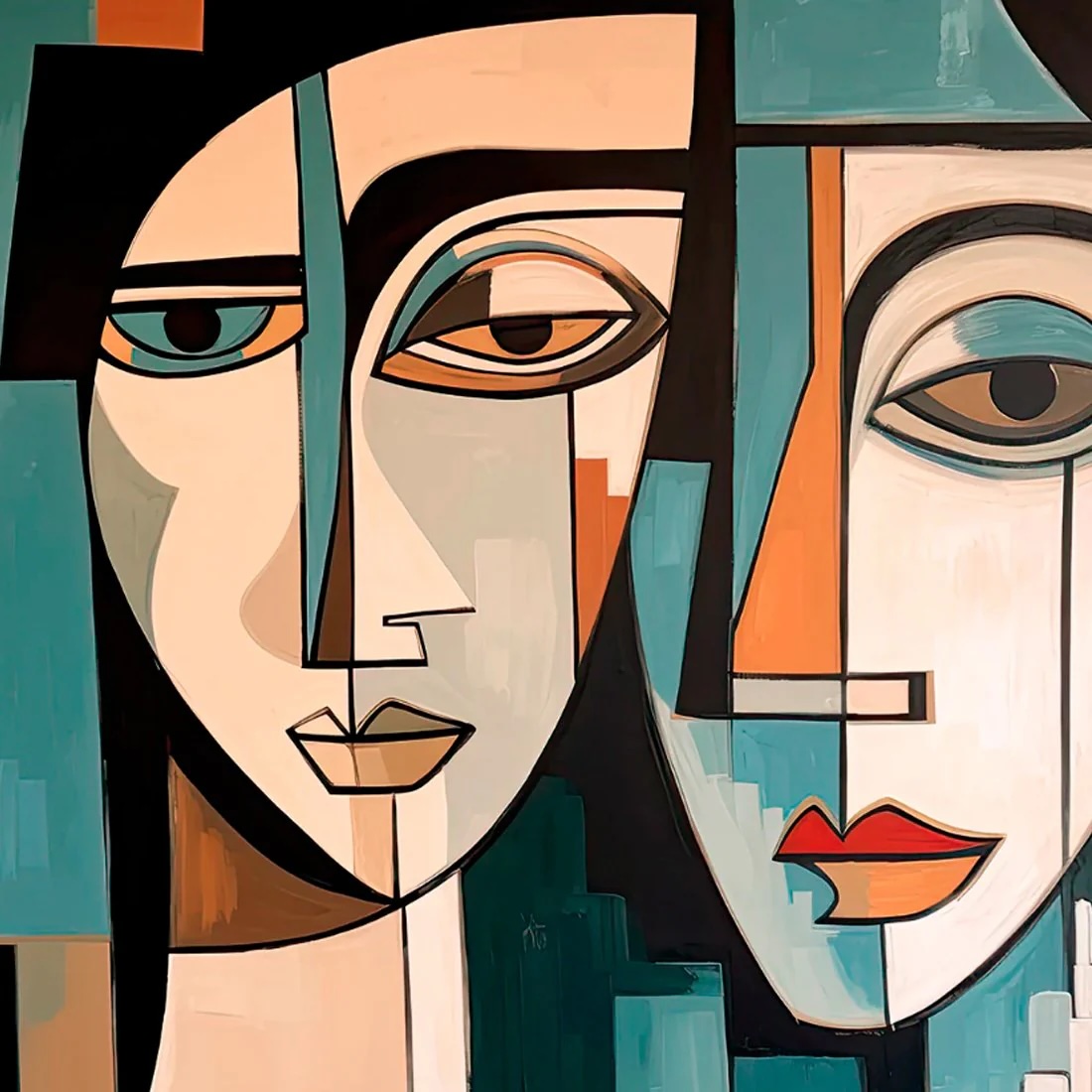}  & Modern cubist portrait style \\
\includegraphics[width=0.5\linewidth]{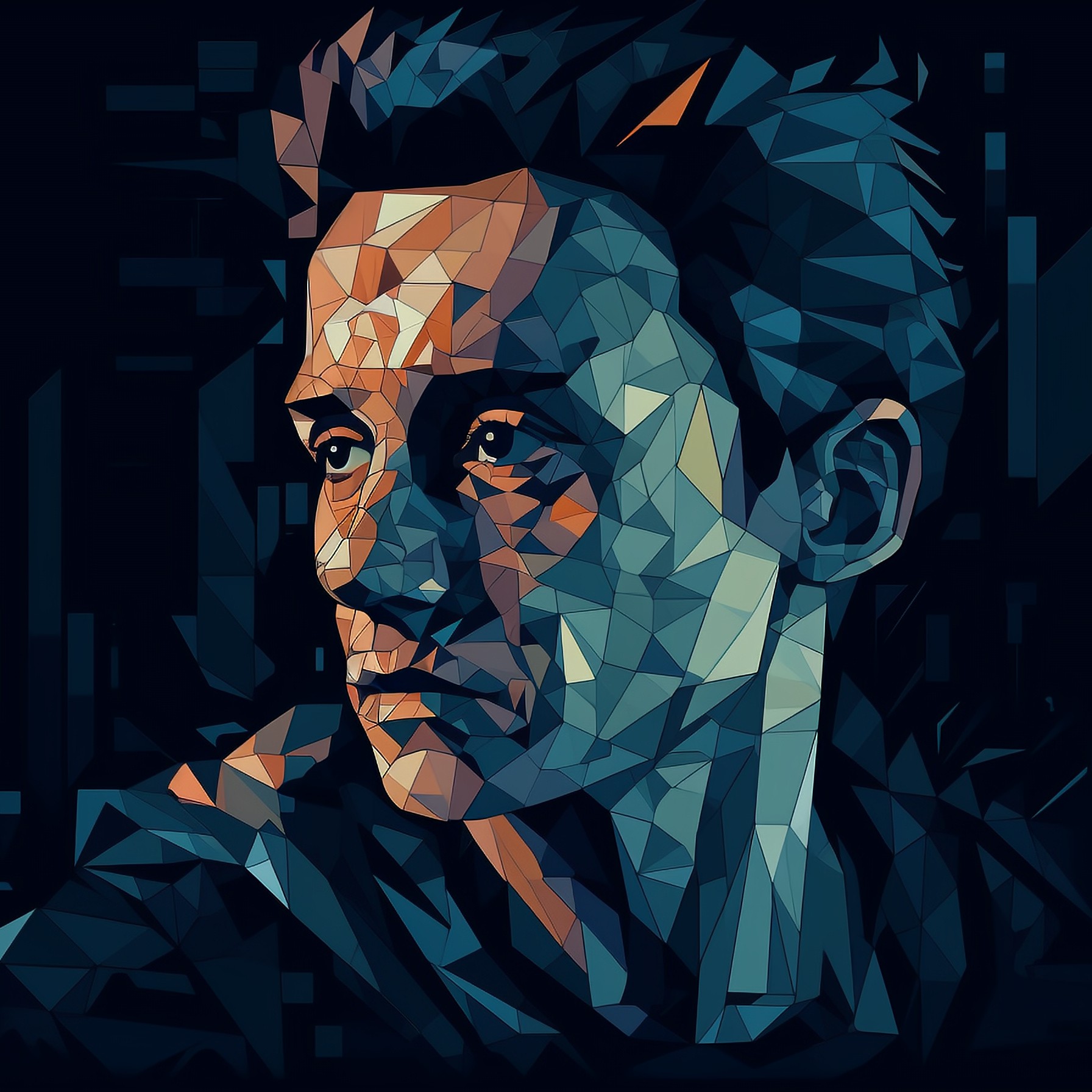} & Low-poly geometric illustration style \\
\includegraphics[width=0.5\linewidth]{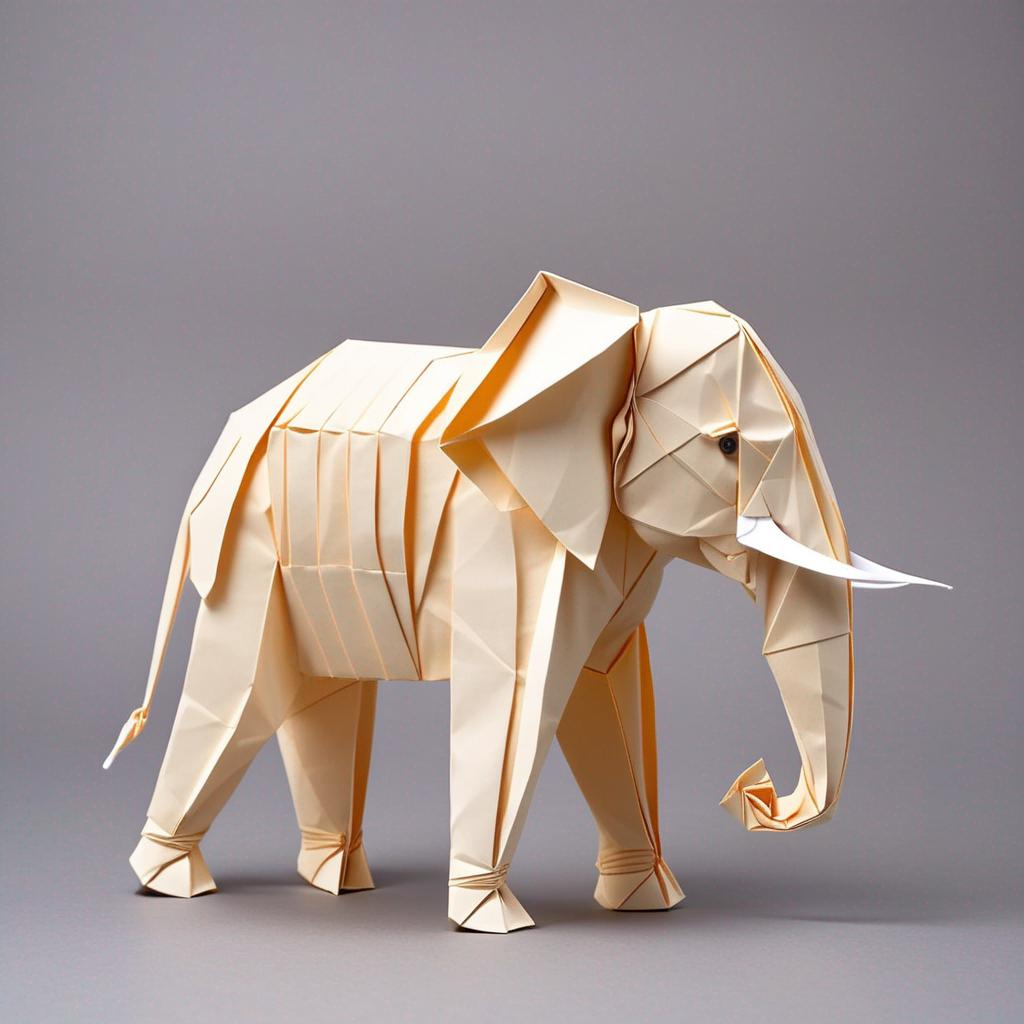} & Origami paper sculpture style \\
\hline
\end{tabular}
\label{tab:style_names}
\end{table}
To compute the HPSv3 score under the specific prompt, we first use a VLM to assign a style name to each style image. The results are shown in Table \ref{tab:style_names}. When using VLMs to compute the aesthetic score, we use the following prompt:

\textit{You are a professional art-style evaluator with strong visual judgment. Your task is to assess the quality of stylized 3D assets generated by different methods.}

\textit{For each evaluation case, I will provide:
\begin{itemize}
    \item 1. A content image, which defines the original object, structure, geometry, and semantic identity.
    \item 2. A style image, which defines the target artistic appearance, including color palette, texture, brushwork, material impression, and overall visual mood.
    \item 3. Rendered images of the generated 3D assets from five different viewpoints.
    \item 4. Results from three different methods.
\end{itemize}}

\textit{You must compare the three generated 3D assets and make a single-choice decision: select the best one among the three methods. Please base your judgment on the following criteria:}

\textit{
\begin{itemize}
    \item 1. Overall visual impression: Evaluate the immediate visual appeal and impact of the asset at first glance. Consider whether the asset looks artistically harmonious, visually coherent, structurally complete, and aesthetically pleasing across different views.
    \item 2. Content and geometry preservation: Compare the stylized 3D asset with the content image. Assess whether the generated asset preserves the original object identity, global structure, local geometry, proportions, and important semantic details.
    \item 3. Style consistency and transfer strength: Compare the stylized 3D asset with the style image. Assess how well the asset reflects the target artistic style, including colors, textures, patterns, material appearance, brush strokes, and overall artistic atmosphere. Also consider whether the style transfer is sufficiently strong without destroying the content structure.
    \item 4. Realism and rendering quality: Evaluate whether the generated 3D asset looks natural, realistic, and high-quality. Consider the quality of materials, surface details, lighting consistency, absence of artifacts, and whether the asset remains visually plausible across the five rendered viewpoints.
\end{itemize}
}

\textit{After considering all four criteria, choose the single best result among the three methods. Your decision should be based only on your visual and artistic perception. You do not need to rely on deeper technical knowledge about 3D generation, rendering, or style transfer methods.}

\textit{Please output only the final choice in the following format: Best Method: [A/B/C].}

\looseness -1
For the user preference study, we used the same evaluation criteria and instructions as those provided to the VLM. The study was conducted in the form of a questionnaire and posed no risk to participants.

\section{Additional Implementation Details}
\label{Appendix B}
\textbf{Implementation for 2D LDM.} In the attention distillation module, self-attention features are extracted from the 10-th and 16-th layers of Stable Diffusion. To better leverage the attention features of Stable Diffusion and align with the diffusion mechanism, we first encode all input images into latent representations using the VAE encoder, and then add random noise corresponding to 200 diffusion steps to these latents by following this equation:
\begin{equation}
z_t = \sqrt{\bar{\alpha}_t}\, z_0 
+ \sqrt{1-\bar{\alpha}_t}\,\epsilon,
\end{equation}
where $z_0$ denotes the original clean latent representation, $z_t$ denotes the noisy latent representation at diffusion timestep $t$, $\epsilon$ is Gaussian noise sampled from the standard normal distribution $\mathcal{N}(0,\mathbf{I})$, and $\bar{\alpha}_t$ is the cumulative noise schedule coefficient defined as
\begin{equation}
\bar{\alpha}_t = \prod_{s=1}^{t} \alpha_s,
\qquad
\alpha_s = 1 - \beta_s,
\end{equation}
where $\beta_s$ controls the noise variance added at timestep $s$.

\textbf{Implementation for Rendered Images.} At each optimization step, we render the 3D asset in 3DGS format into images, which are then fed into the 2D LDM. First, following Trellis, we define a camera trajectory using the following equation:
\begin{equation}
\begin{aligned}
\theta_i &= \frac{2\pi i}{N-1}, \\
\phi_i &= 0.25 + 0.5 \sin\!\left(\frac{2\pi i}{N-1}\right), \\
\rho_i &= 0,
\qquad i \in \{0,\dots,N-1\}.
\end{aligned}
\end{equation}

\begin{equation}
\mathbf{c}_i
=
r
\begin{bmatrix}
\sin(\theta_i)\cos(\phi_i) \\
\cos(\theta_i)\cos(\phi_i) \\
\sin(\phi_i)
\end{bmatrix},
\qquad r=2.
\end{equation}
where $N$ denotes the total number of rendered views, and $i$ is the view index. 
$\theta_i$, $\phi_i$, and $\rho_i$ denote the yaw, pitch, and roll angles of the $i$-th camera view, respectively. 
The yaw angle $\theta_i$ changes linearly from $0$ to $2\pi$, forming a full circular trajectory around the object. 
The pitch angle $\phi_i$ varies sinusoidally, allowing the camera to observe the object from smoothly changing elevations, while the roll angle $\rho_i$ is fixed to zero for all views. 
The camera center $\mathbf{c}_i$ is placed on a sphere with radius $r=2$ around the origin. We render images from 300 viewpoints and then uniformly select 5 views to feed into the 2D LDM, reducing memory consumption and improving optimization efficiency. Moreover, the resolution of all rendered images is 512 $\times$ 512.

\textbf{Implementation for Overall Optimization.} Both the Floater Removal loss and the Color Speckle Suppression loss impose a certain degree of smoothing on the 3D latents, which may hinder style transfer. To avoid interfering with the early guidance of attention distillation on the 3D latents, we introduce these two losses only after 400 sampling steps.

\section{Additional Results.}
\label{Appendix C}
\subsection{Supplementary Qualitative and Quantitative Results}
Since our motivation is to enrich the diversity of 3D asset generation through 3D style transfer, and since 3D assets are primarily used in applications such as gaming and animation, we mainly select content images with artistic styles as generation conditions in the main paper. In this subsection, we select five real-world objects and provide the corresponding style transfer results. Additional experimental results that cannot be included in the main paper due to space limitations are also presented in this subsection. Videos are provided in the supplementary material.
\begin{figure}[H]
    \centering
    \includegraphics[width=\linewidth]{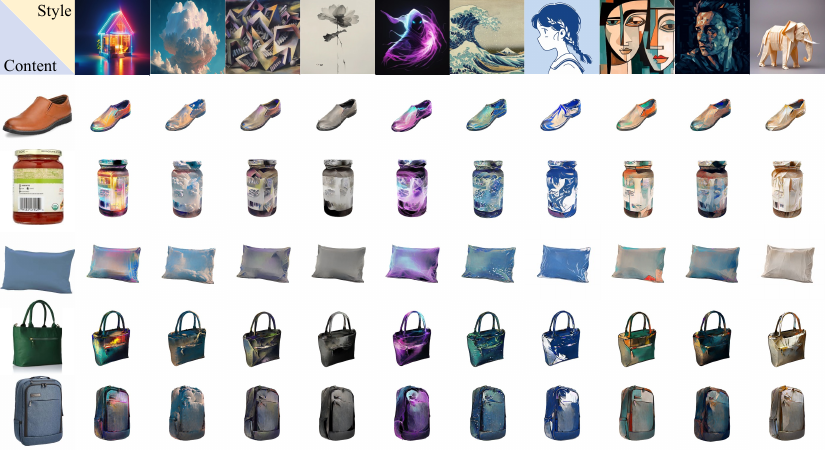}
    \caption{Qualitative results of real-world objects using our method.}
    \label{fig7}
\end{figure}
\begin{table}[H]
\centering
\looseness -1
\caption{Quantitative comparison of real-world objects. The best results are in \textbf{bold}. GP: general prompt, SP: specific prompt, AS denotes aesthetic score (averaged on two VLMs), and UP denotes user preference.}
\label{tab3}
\begin{tabular}{ccccc}
\hline
Method        & HPSv3 (GP)$\uparrow$ & HPSv3 (SP)$\uparrow$ & AS (\%)$\uparrow$ & UP (\%)$\uparrow$ \\ \hline
MorphAny3D    & -5.4793    & -5.7145    & 6.43         & 3.70    \\
StyleSculptor & -2.2589     & -2.4302    & 24.29         & 11.11   \\
DiLAST (Ours) & \textbf{2.7565}     & \textbf{2.3995}     &  \textbf{69.28}       & \textbf{85.19}   \\ \hline
\end{tabular}
\end{table}
\begin{figure}[H]
    \centering
    \includegraphics[width=\linewidth]{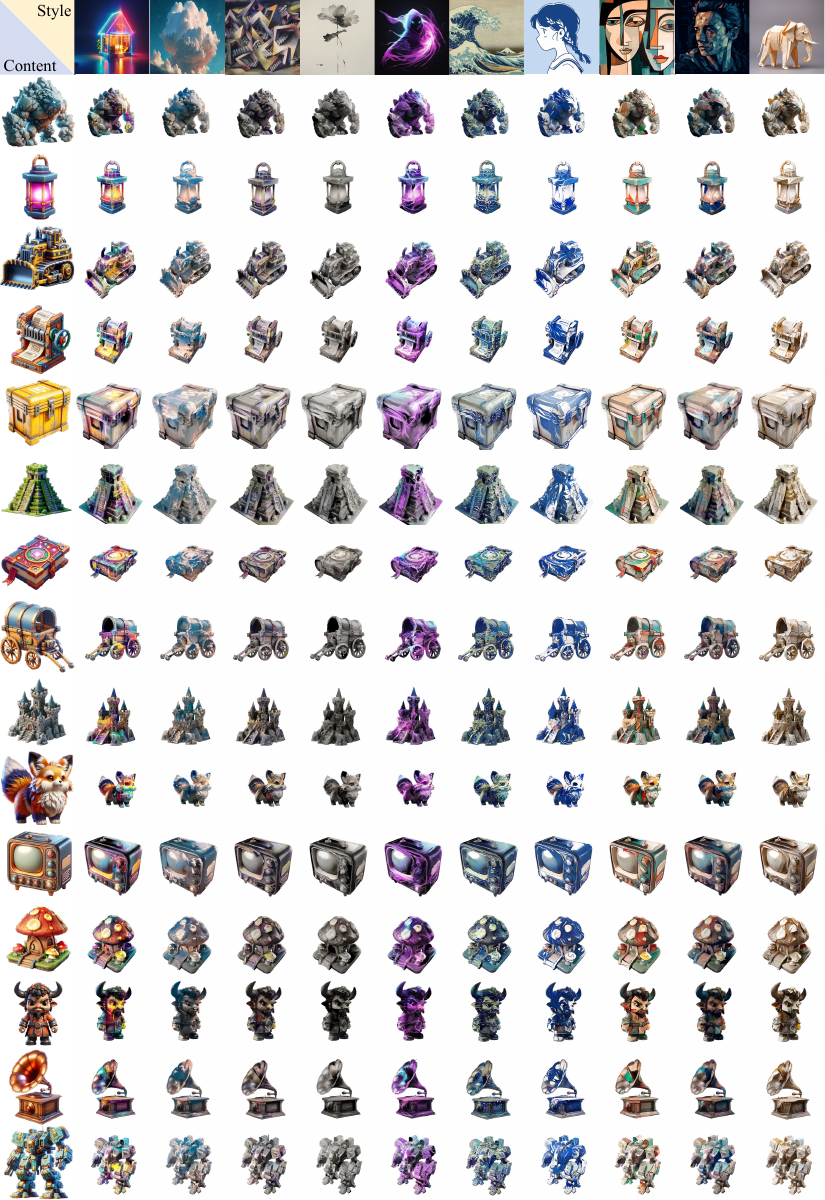}
    \caption{Supplementary qualitative results for the main paper by using DiLAST.}
    \label{fig8}
\end{figure}
\begin{figure}[H]
    \centering
    \includegraphics[width=0.95\linewidth]{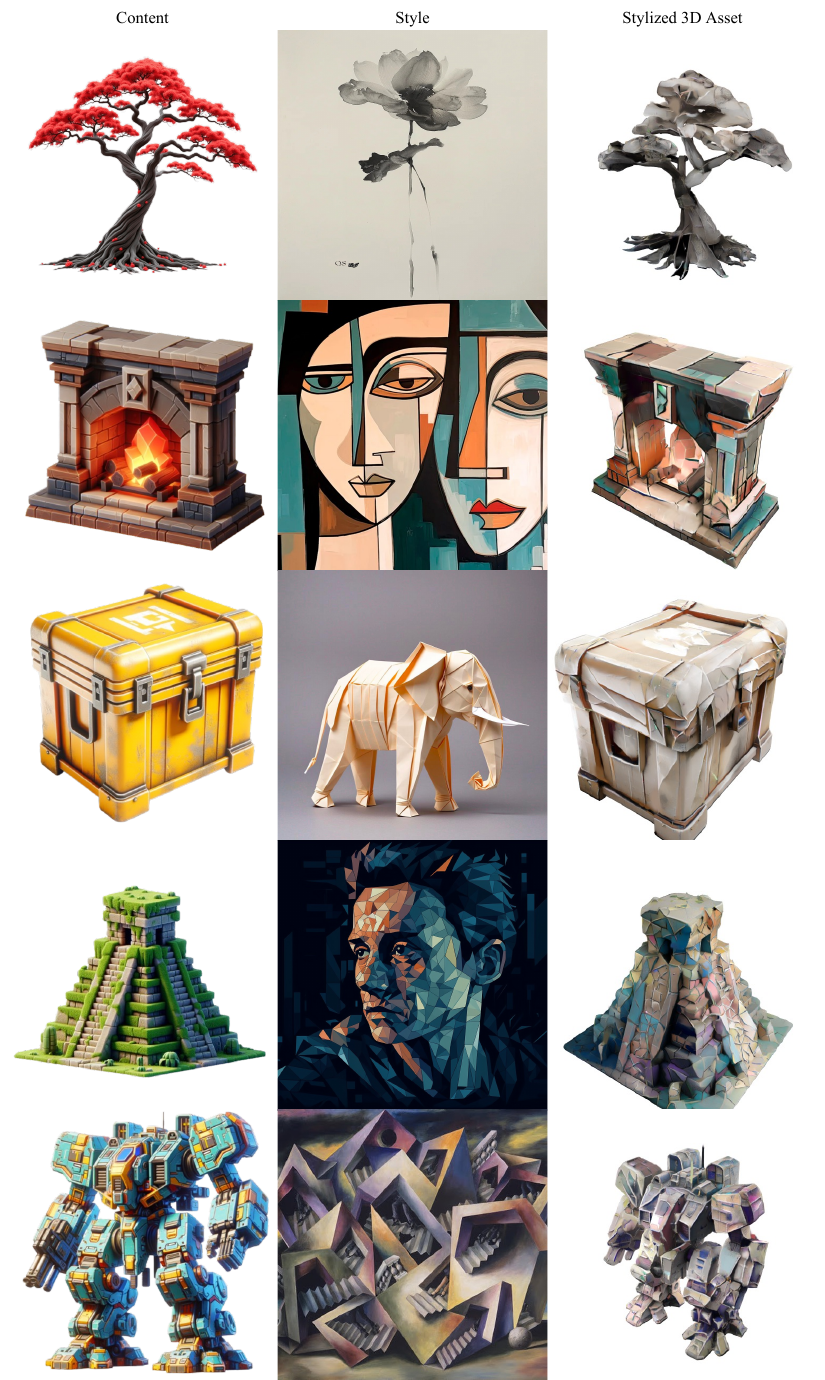}
    \caption{Selected zoomed-in results.}
    \label{fig9}
\end{figure}
\begin{figure}[H]
    \centering
    \includegraphics[width=0.95\linewidth]{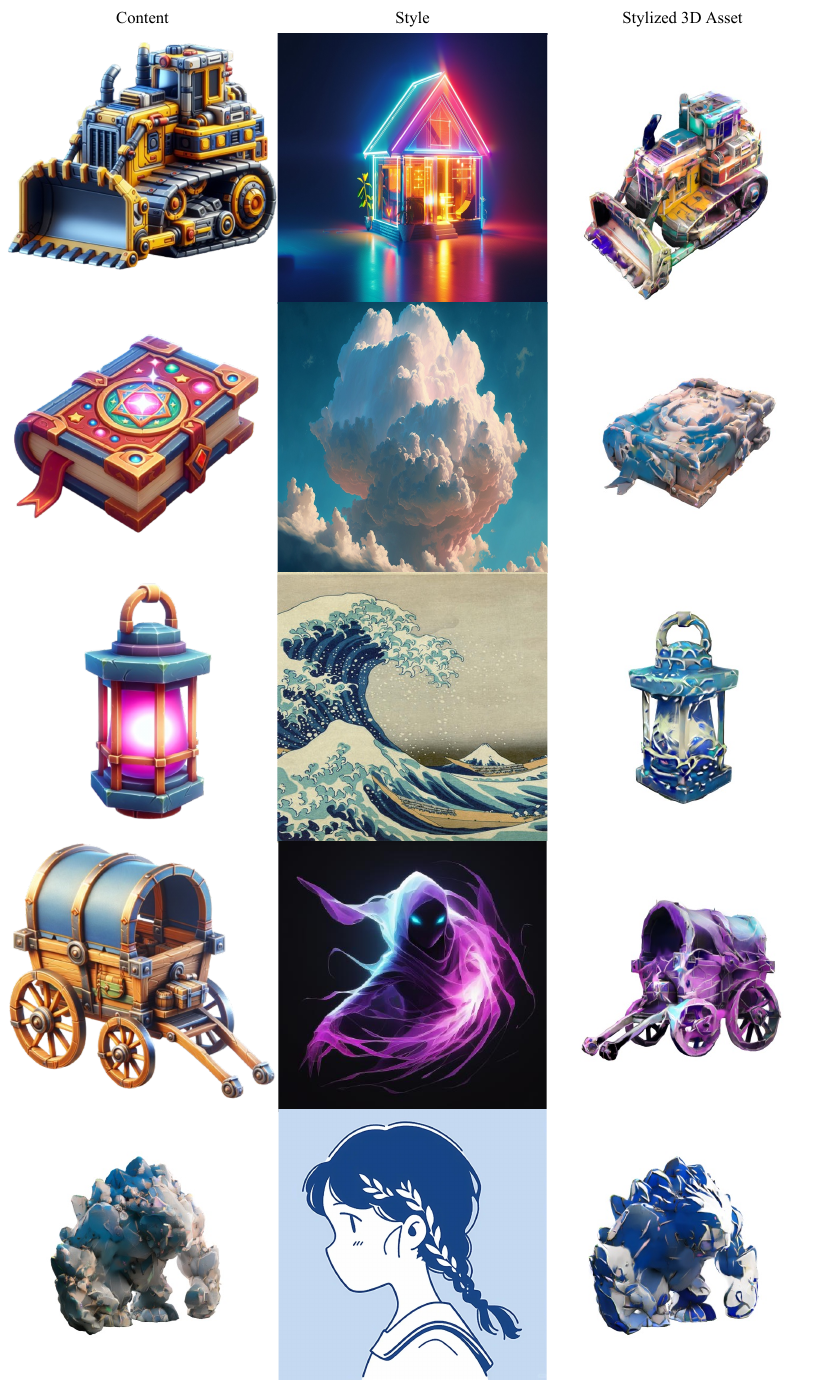}
    \caption{Selected zoomed-in results (part 2).}
    \label{fig10}
\end{figure}
\begin{figure}[H]
    \centering
    \includegraphics[width=0.95\linewidth]{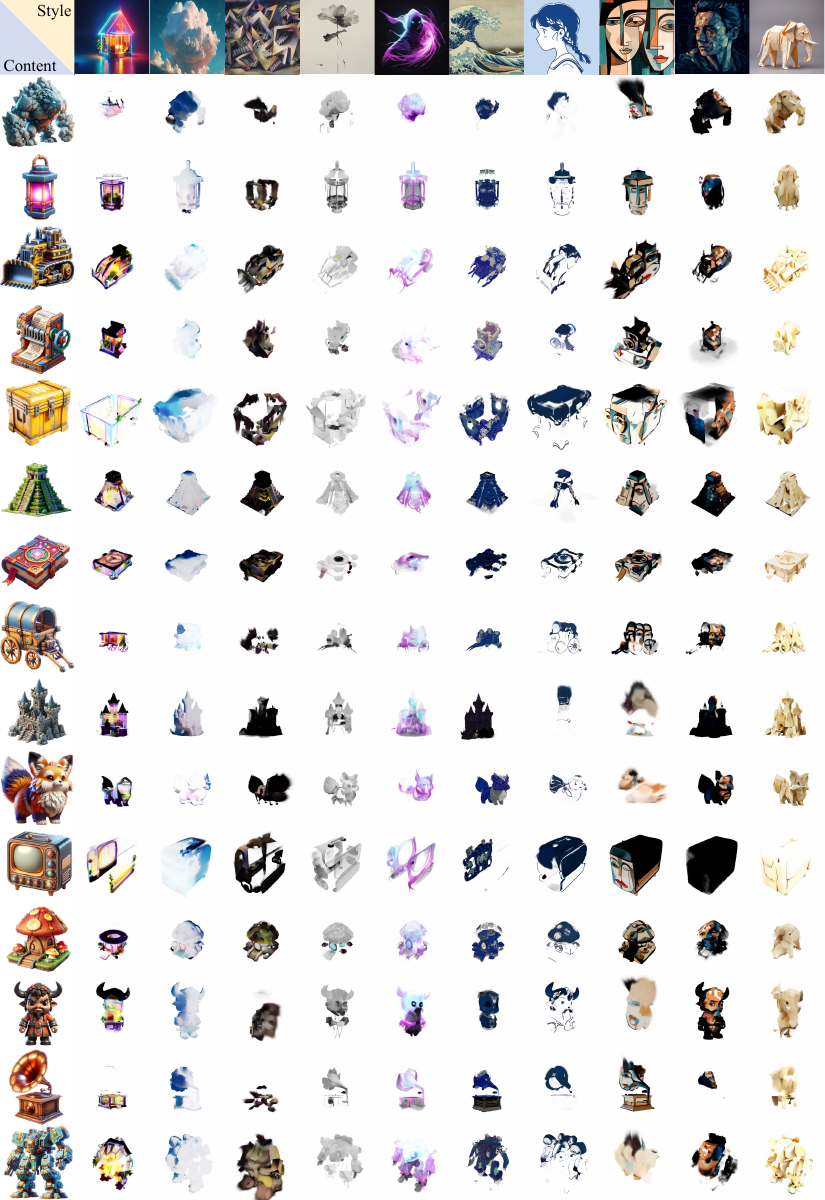}
    \caption{Supplementary qualitative results by using MorphAny3D.}
    \label{fig11}
\end{figure}
\begin{figure}[H]
    \centering
    \includegraphics[width=0.95\linewidth]{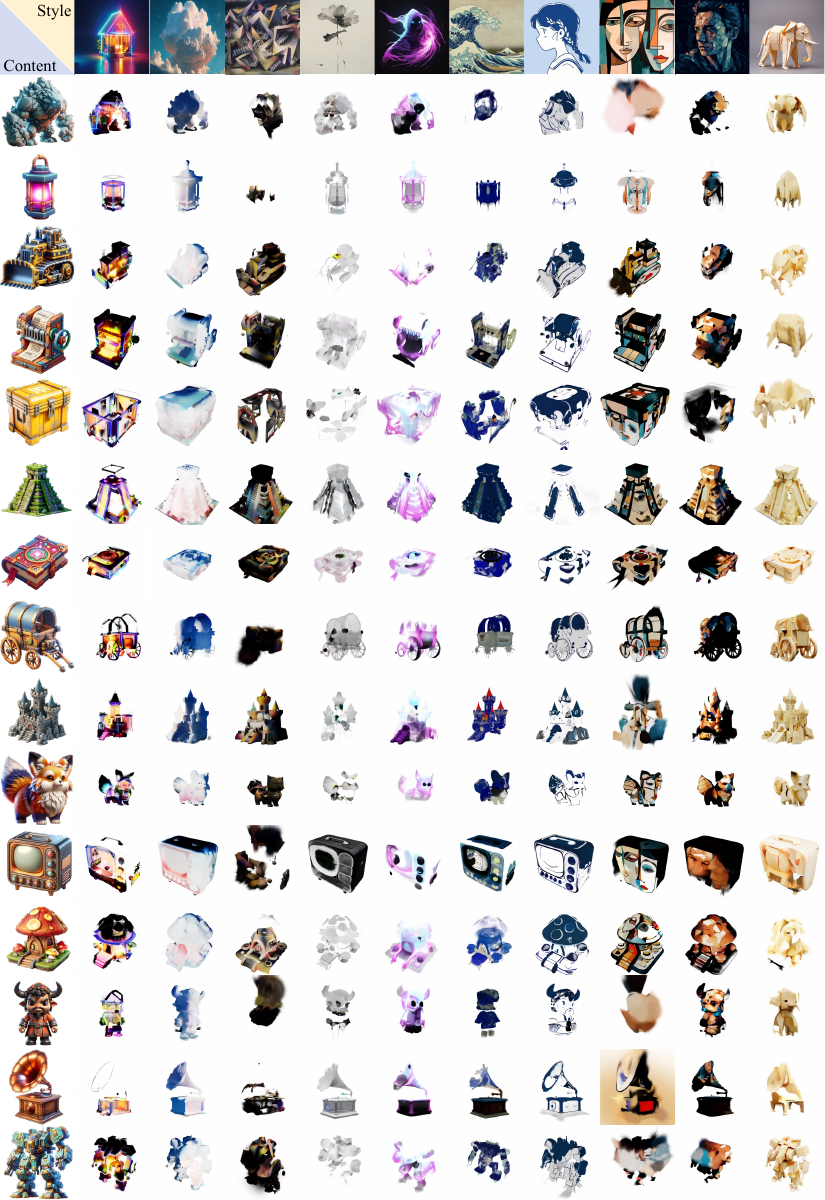}
    \caption{Supplementary qualitative results by using StyleSculptor.}
    \label{fig12}
\end{figure}
Figure \ref{fig7} and Table \ref{tab3} present the qualitative and quantitative results of the generated 3D assets for real-world objects, respectively. Figures \ref{fig8}, \ref{fig11}, and \ref{fig12} show additional stylized 3D assets generated by DiLAST, MorphAny3D, and StyleSculptor, respectively. To better showcase the details of the stylized 3D assets generated by our method, we select several examples and provide zoomed-in visualizations in Figures \ref{fig9} and \ref{fig10}.

\subsection{Analysis of Transfer-then-Generate Approaches}
\begin{figure}[H]
    \centering
    \includegraphics[width=0.95\linewidth]{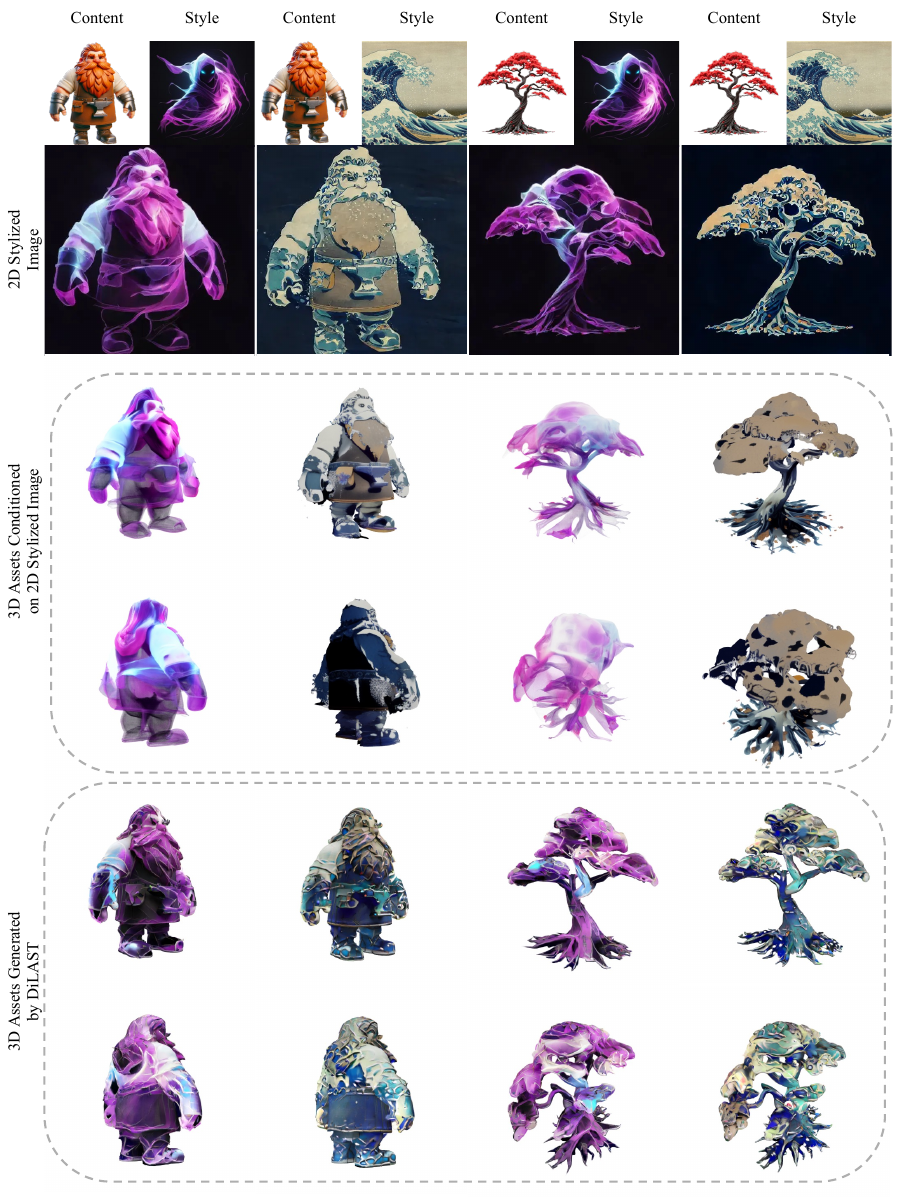}
    \caption{Qualitative comparison between TTG approach and DiLAST.}
    \label{fig13}
\end{figure}

Transfer-then-generate (TTG) approaches first apply a 2D style transfer method to obtain a stylized 2D image, and then use this image as the condition for 3D asset generation. StyleSculptor \cite{qu2025stylesculptor} has pointed out that such approaches can lead to distortions in 3D geometry and even introduce semantic inaccuracies. In this section, we further illustrate this issue through several examples. As shown in Figure \ref{fig13}, stylized 3D assets generated by transfer-then-generate methods suffer from artifacts, missing details, and inconsistencies with the conditioning images. In contrast, DiLAST avoids these issues and produces more faithful stylized 3D assets. This is mainly because some 2D stylized images serve as out-of-distribution conditions for 3D generative models.
\begin{figure}[H]
    \centering
    \includegraphics[width=0.9\linewidth]{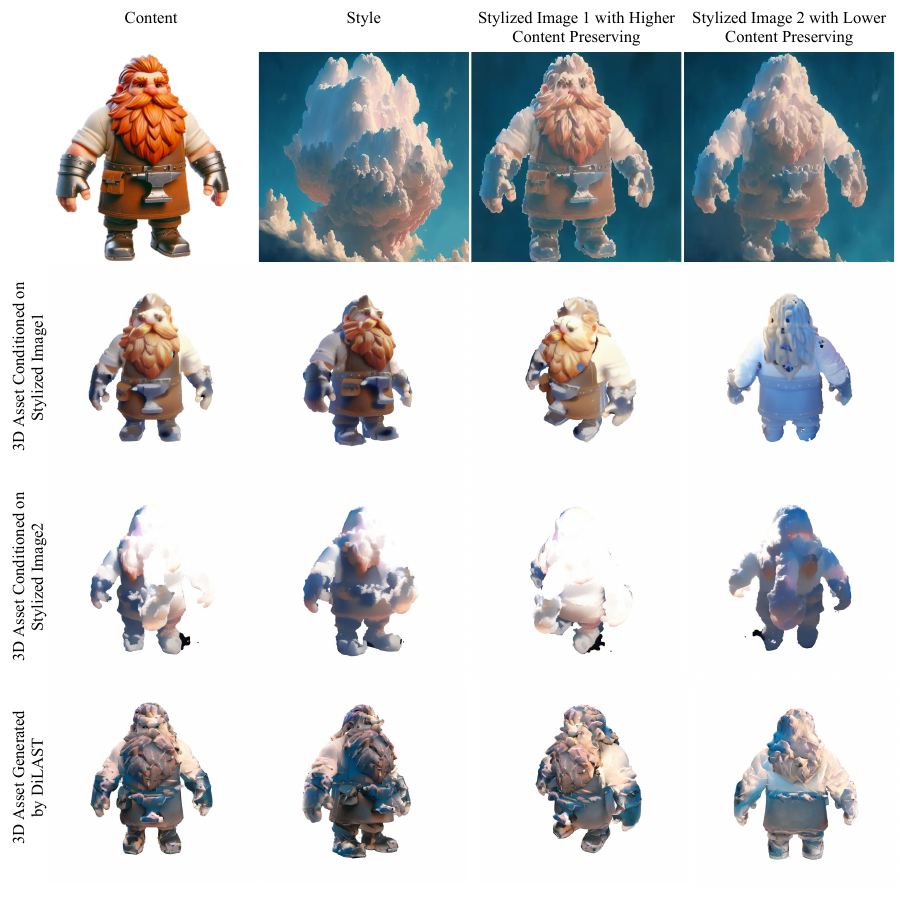}
    \looseness -1
    \caption{Qualitative comparison between different TTG approach settings and DiLAST.}
    \vspace{-3mm}
    \label{fig14}
\end{figure}
Another issue with TTG is that, regardless of whether the content preservation weight is set high or low during 2D stylization, the resulting 3D asset often fails to reflect the target style faithfully. As shown in Figure \ref{fig14}, when content preservation is too strong, the style cannot be effectively transferred. Conversely, when content preservation is too weak, the stylized image deviates further from the training distribution of the 3D generative model, leading to poor generation quality. In practice, it is often difficult for 2D style transfer to find a suitable level of content preservation that remains "recognizable" to the 3D model. In contrast, our method directly guides the 3D latents, indirectly injecting an appropriate degree of style into the latent space while achieving suitable content preservation, without being affected by the 2D stylized image.

\vspace{-3mm}
\section{Limitations}
\vspace{-3mm}

\label{Appendix D}
Our method and the quality of style transfer depend on the pretrained 2D and 3D generative models. When memory permits, stronger 2D LDMs, such as SDXL, can be used to provide more robust guidance. In addition, we expect that future advances in more powerful 3D generative models will further improve the performance of our method.

\vspace{-3mm}
\section{Broader Impacts}
\vspace{-3mm}
\label{Appendix E}
\looseness -1
DiLAST can lower the barrier for creating stylized 3D assets, benefiting applications such as game development, animation, virtual reality, digital art, and personalized content creation. By enabling flexible style transfer without retraining 3D generative models, it may reduce manual modeling effort and make 3D creation more accessible. However, the method could also be misused to imitate copyrighted artistic styles, create unauthorized derivative assets, or generate misleading visual content. We encourage responsible use, including respecting intellectual property rights, obtaining proper permissions for style references, and clearly disclosing AI-generated or AI-edited 3D content when appropriate.



\end{document}